\documentclass[runningheads]{llncs}
\usepackage{graphicx}
\usepackage{amsfonts}
\usepackage{amssymb}
\usepackage{siunitx}
\usepackage{amsbsy}
\usepackage{cite}
\usepackage{subfig}
\usepackage{color}

\begin{document}
\title{BERTSurv: BERT-Based Survival Models for Predicting Outcomes of Trauma Patients}
\titlerunning{BERTSurv}
%
\author{Yun Zhao\inst{1} \and
Qinghang Hong\inst{1} \and
Xinlu Zhang\inst{1} \and
Yu Deng\inst{2} \and
Yuqing Wang\inst{1} \and
Linda Petzold\inst{1}}

\authorrunning{Y. Zhao et al.}
%
\institute{Department of Computer Science, University of California, Santa Barbara 
\and
Feinberg School of Medicine, Northwestern University
\email{\\yunzhao@cs.ucsb.edu}\\
}
\maketitle              
\begin{abstract}
\begin{quote}
Survival analysis is a technique to predict the times of specific outcomes, and is widely used in predicting the outcomes for  intensive care unit (ICU) trauma patients. Recently, deep learning models have drawn increasing attention in healthcare. However, there is a lack of deep learning methods that can model the relationship between measurements, clinical notes and mortality outcomes. In this paper we introduce BERTSurv, a deep learning survival framework which applies Bidirectional Encoder Representations from Transformers (BERT) as a language representation model on unstructured clinical notes, for mortality prediction and survival analysis. We also incorporate clinical measurements in BERTSurv. With binary cross-entropy (BCE) loss, BERTSurv can predict mortality as a binary outcome (mortality prediction). With partial log-likelihood (PLL) loss, BERTSurv predicts the probability of mortality as a time-to-event outcome (survival analysis). We apply BERTSurv on Medical Information Mart for Intensive Care III (MIMIC III) trauma patient data. For mortality prediction, BERTSurv obtained an area under the curve of receiver operating characteristic curve (AUC-ROC) of 0.86, which is an improvement of 3.6\% over baseline of multilayer perceptron (MLP) without notes. For survival analysis, BERTSurv achieved a concordance index (C-index) of 0.7. In addition, visualizations of BERT's attention heads help to extract patterns in clinical notes and improve model interpretability by showing how the model assigns weights to different inputs.

\end{quote}
\end{abstract}

\section{Introduction}

Trauma is the leading cause of death from age 1 to 44. More than 180,000 deaths from trauma occur each year in the United States~\cite{Trauma}. Most trauma patients die or are discharged quickly after being admitted to the ICU. Care in the first few hours after admission is critical to patient outcome, yet this time period is more prone to medical decision errors in ICUs~\cite{ICU} than later periods. Therefore, early and accurate prediction for trauma patient outcomes is essential for ICU decision making.

Medical practitioners use survival models to predict the outcomes for trauma patients~\cite{YY_SVTM}. Survival analysis is a technique to model the distribution of the outcome time. The Cox model~\cite{Cox} is one of the most widely used survival models with linear proportional hazards. Faraggi-Simon's network~\cite{Faraggi} is an extension of the Cox model to nonlinear proportional hazards using a neural network. DeepSurv~\cite{DeepSurv} models interactions between a patient’s covariates and treatment effectiveness with a Cox proportional hazards deep neural network. However, these existing models deal only with well-structured measurements and do not incorporate information from unstructured clinical notes, which can offer significant insight into patients' conditions.

The transformer architecture has taken over sequence transduction tasks in natural language processing (NLP). Transformer is a sequence model that adopts a fully attention-based approach instead of traditional recurrent architectures. Based on Transformer, BERT~\cite{Bert} was proposed for language representation and achieved state-of-the-art performance on many NLP tasks. There has also been increasing interest in applying deep learning to end-to-end e-health data analysis~\cite{deep_ICU}. Biobert~\cite{biobert} extends BERT to model biomedical language representation. Med-BERT~\cite{medbert} modifies BERT by leveraging domain specific hierarchical code embedding and layer representation to generate sequential relationships in the clinical domain. G-BERT~\cite{gbert} combines Graph Neural Networks (GNNs) and BERT for medical code
representation and medication recommendation. Clinical BERT~\cite{clinical_BERT1, clinical_BERT2} explores and pre-trains BERT using clinical notes. Clearly, there is an unmet need to include unstructured text information in deep learning survival models for patient outcome predictions.

In this paper we propose BERTSurv, a deep learning survival framework for trauma patients which incorporates clinical notes and measurements for outcome prediction. BERTSurv allows for both mortality prediction and survival analysis by using BCE and PLL loss, respectively. Our experimental results indicate that BERTSurv can achieve an AUC-ROC of 0.86, which is an improvement of 3.6\% over the baseline of MLP without notes on mortality prediction.
 
The key contributions of this paper are:

1. We propose BERTSurv: a BERT-based deep learning framework to predict the risk of death for trauma patients. To the best of our knowledge, this is the first paper applying BERT on unstructured text data combined with measurements for survival analysis.

2. We evaluate BERTSurv on the trauma patients in MIMIC III. For mortality prediction, BERTSurv achieves an AUC-ROC of 0.86, which outperforms baseline of  MLP without notes by 3.6\%. For survival analysis, BERTSurv achieved a C-index of 0.7 on trauma patients, which outperforms a Cox model with a C-index of 0.68.

3. We extract patterns in clinical notes by performing attention mechanism visualization, which improves model interpretability by showing how the model assigns weights to different clinical input texts with respect to survival outcomes.

This paper is organized as follows: Section~\ref{Data} describes how we processed the MIMIC trauma dataset. We present BERTSurv in Section~\ref{Model} and describe the background of BERT and survival analysis in Section~\ref{BERT} and Section~\ref{Survival_Analysis}. Evaluation and discussion are given in Sections~\ref{ex} and~\ref{Dis}, respectively.

\section{Dataset}\label{Data}

BERTSurv is applied to the data from trauma patients selected using the ICD-9 code from the publicly available MIMIC III dataset~\cite{MIMIC}, which provides extensive electronic medical records for ICU admissions at the Beth Israel Deaconess Medical Center between 2001 and 2012. The measurements, clinical notes, expire flag (0 for discharge and 1 for death), and death/discharge time for each patient were used to train and test BERTSurv. The patient data were aggregated over the first 4 hours to obtain the initial state of each individual admission. We took the average for each of the measurements taken during this time period, and concatenated all of the clinical notes together. Considering the missing value issue and redundancy in MIMIC III, we selected 21 common features as our representative set: blood pressure, temperature, respiratory rate, arterial PaO2, hematocrit, WBC, creatinine, chloride, lactic acid, BUN, sodium (Na), glucose, PaCO2, pH, GCS, 
heart rate, FiO2, potassium, calcium, PTT and INR. Our feature set overlaps 65\% of the measurements required by APACHE III~\cite{apache}. We also extracted 4 demographic predictors: weight, gender, ethnicity and age. 

As is common in medical data, MIMIC III contains many missing values in the measurements, and the notes are not well-formatted. Thus, data preprocessing is very important to predict outcomes. To deal with the missing data issue, we first removed patients who have a missing value proportion greater than 0.4 and then applied MICE~\cite{MICE} data imputation for the remainder of the missing values. For the clinical notes, we removed formatting, punctuation, non-punctuation symbols and stop words. In addition to the most commonly used English stop words, our stop word dictionary includes a few specific clinical and trauma related stop words: \textit{doctor, nurse and measurement}, etc.
Following this preprocessing, our trauma dataset includes 1860 ICU patients, with 21 endogenous measurements, 4 exogenous measurements and notes. The sample class ratio between class $0$ (discharge) and class $1$ (death) is $1206 : 654$.

\section{Methods}\label{Methods}
In this section we first describe the framework of BERTSurv. Then we introduce some basics of BERT and survival analysis, which are the key components of BERTSurv.

\subsection{BERTSurv}\label{Model}

\begin{figure}
\centering
\includegraphics[width=\textwidth]{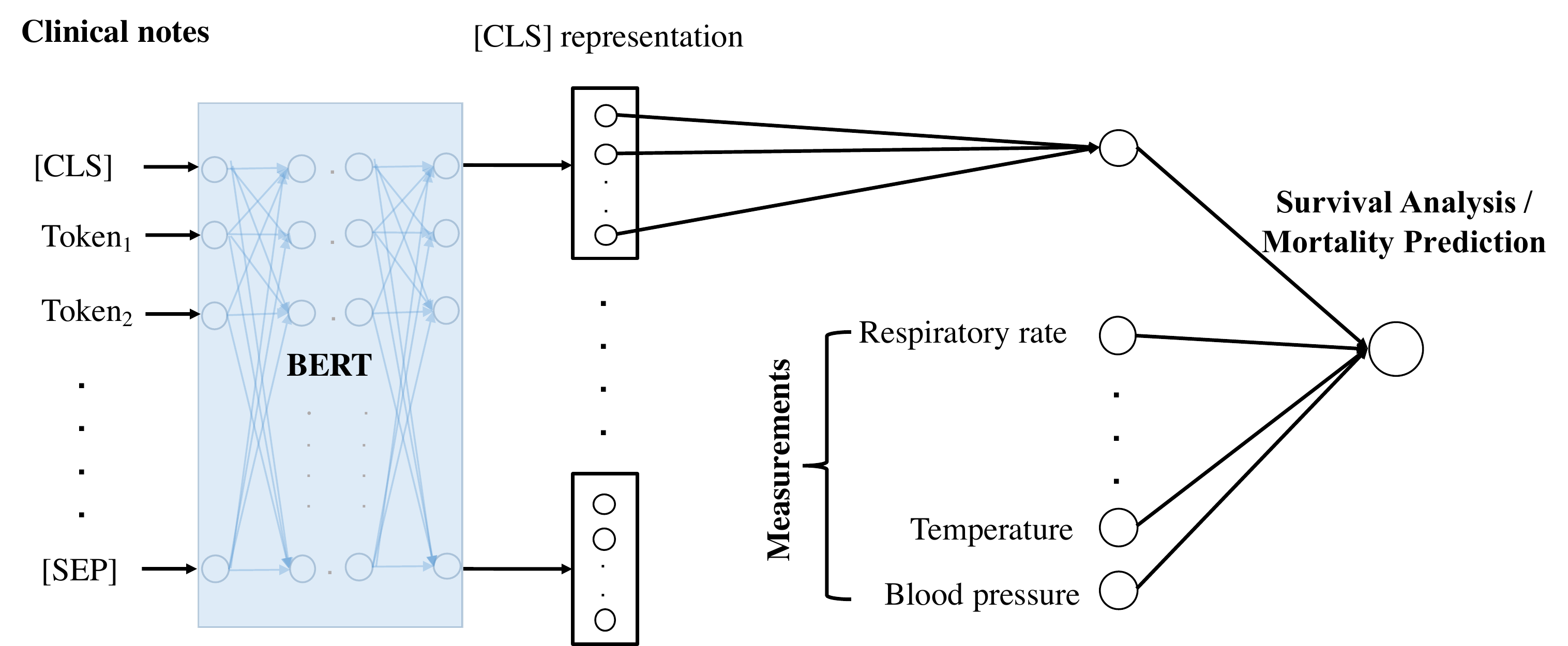}
\caption{The framework of BERTSurv. [CLS] is a special
symbol added in front of every clinical note sample, and [SEP] stands for a special separator token. BERTSurv consists of three main parts: BERT, measurements and output layer for mortality prediction or survival analysis. First, we input a set of diagnostics and nurse notes to BERT pretrained on masked language modeling and next sentence prediction. The [CLS] representation, is
treated as the representation of the input notes. Then we concatenate the [CLS] representation and measurements as input and fine-tune
BERTSurv for downstream survival analysis.  
} \label{fig_model}
\end{figure}

Our model architecture, shown in Fig~\ref{fig_model}, consists of BERT embedding of clinical notes concatenated with measurements followed by feed forward layers. The output for BERTSurv is a single node $h_{\theta}(x_i)$ parameterized by the weights of the neural network $\theta$, which estimates either the probability of mortality or the hazard risk. For mortality prediction, we apply BCE loss to predict outcomes of death or discharge:
\begin{equation}
\label{eq:bce}
{\rm BCE Loss} := \sum\limits_{i=1}^np(y_i) \log(h_{\theta}(x_i)),
\end{equation}
where $x_i$ and $y_i$ represent inputs and outcomes for the $i$th patient, respectively.

To estimate $\theta$ in survival analysis, similar to the Faraggi-Simon network~\cite{Faraggi,DeepSurv}, we minimize the PLL loss function, which is the average negative log partial likelihood:

\begin{equation}
\label{eq:pll}
{\rm{PLL Loss}} := -\frac{1}{N_{D = 1}}\sum\limits_{i:D_i = 1}(h_{\theta}(x_i) - \log\sum\limits_{j\in R(T_i)}\exp(h_{\theta}(x_j))),
\end{equation}
where $N_{D = 1}$ is the number of patients with an observable death. The risk set $R_i = \{j : T_j \geq T_i\}$ is the set of those patients under risk at $T_i$.

We use batch normalization through normalization of the input layer by re-centering and re-scaling~\cite{BatchNorm}. We apply rectified linear unit(ReLU) or scaled exponential linear units (SELU) as the activation function. For regularization, dropout~\cite{Dropout} is implemented to avoid overfitting. Dropout prevents co-adaptation of hidden units by randomly dropping out a proportion of the hidden units during backpropagation. BCE/PLL loss is minimized with the Adam optimizer~\cite{Adam} for training. 

BERTSurv is implemented in Pytorch~\cite{Pytorch}.
We use a 
Dell 32GB NVIDIA Tesla M10 Graphics Card
GPU (and significant CPU power and memory for pre-processing tasks) for training, validation and testing. The hyperparameters of the network include: BERT choice ($\rm BERT_{BASE}$ or clinical BERT~\cite{clinical_BERT1}), sequence length, batch size, learning rate, dropout rate, training epochs and activation function (ReLU or SELU).

\subsection{BERT}\label{BERT}
A key component of BERTSurv is the BERT language representation model. BERT is a Transformer-based language representation model, which is designed to pre-train deep bidirectional representations from unlabeled text by jointly considering context from both directions (left and right). Using BERT, the input representation for each token in the clinical notes is the sum of the corresponding token embeddings, segmentation embeddings and position embeddings. WordPiece token embeddings~\cite{WordPiece} with a 30,000 token vocabulary are applied as input to BERT.
The segment embeddings identify which sentence the token is associated with. The position embeddings of a token are a learned set of parameters corresponding to the token’s position in the input sequence. An attention function maps a query and a set of key-value pairs to an output. The attention function takes a set of queries $Q$, keys $K$, and values $V$ as inputs and is computed on an input sequence using the embeddings associated with the input tokens. To construct $Q$, $K$ and $V$, every input embedding is multiplied by the learned sets of weights. The attention function is

\begin{equation}
\label{eq:attention}
{\rm Attention}(Q, K, V) = {\rm softmax}(\frac{QK^T}{\sqrt{d_k}}V),
\end{equation}
where $d_k$ is the dimensionality of $Q$ and $K$. The dimension of $V$ is $d_v$.
A multi-head attention mechanism allows BERT to jointly deal with information from different representation subspaces at different positions with several ($h$) attention layers running in parallel:

\begin{equation}
\label{eq:multi-head-attention}
{\rm MultiHead}(Q, K, V ) = {\rm Concat(head_1, ..., head_h)}W^O,
\end{equation}
where ${\rm head_i = Attention}(QW_i^Q, KW_i^K, VW_i^V)$. Parameter matrices $W_i^Q \in \mathbb{R}^{d_{\rm model} \times {d_k}}$,  $W_i^K \in \mathbb{R}^{d_{\rm model} \times {d_k}}$, $W_i^V \in \mathbb{R}^{d_{\rm model} \times {d_v}}$ and $W^O \in \mathbb{R}^{h {d_v} \times d_{\rm model}}$ are the learned linear projections from $Q$, $K$, $V$ to $d_k$, $d_k$ and $d_v$ dimensions.

In BERTSurv, we use pretrained BERT of $\rm BERT_{BASE}$ and clinical BERT~\cite{clinical_BERT1} for clinical note embedding, and focus on fine-tuning for survival analysis. 

\subsection{Survival Analysis}\label{Survival_Analysis}

Another key component of BERTSurv is survival analysis. Survival analysis~\cite{klein2005survival, kalbfleisch2011statistical}
is a statistical methodology for analyzing the expected duration until one or more events occur. The survival function $S(t)$, defined as $S(t) = P(T \geq t)$, gives the probability that the time to the event occurs later than a given time $t$. The cumulative distribution function (CDF) of the time to event gives the cumulative probability for a given t-value:
\begin{equation}
F(t) = P(T < t) = 1-S(t).
\label{cdf}
\end{equation}

The hazard function $h(t)$ models the probability that
an event will occur in the time interval $[t, t+\Delta t)$ given that the event has not occurred before:
\begin{equation}
h(t)=\lim_{\Delta t\rightarrow 0}\frac{P(t\leq T<t+\Delta t\mid T\geq t)}{\Delta t}=\frac{f(t)}{S(t)},
\label{hazard}
\end{equation}
where $f(t)$ is the probability density function (PDF) of the time to event. A greater hazard
implies a greater probability of event occurrence. Note from Equ.~\ref{cdf} that $-f(t)$ is the derivative of $S(t)$. Thus Equ.~\ref{hazard} can be rewritten as
\begin{equation}
h(t)= - \frac{dS(t)}{dt}* \frac{1}{S(t)} = -\frac{d}{dt}\log(S(t)).
\label{ode}
\end{equation}
By solving Equ.~\ref{ode} and introducing the boundary condition $S(0)=1$ (the event can not occur before duration 0), the relationship between $S(t)$ and $h(t)$ is given by
\begin{equation}
\label{eq:stht}
S(t) = \exp\left ( -\int_{0}^{t}h(s) \mathrm{d}s \right ).
\end{equation}

The Cox model~\cite{Cox} is a well-recognized survival model.
It defines the hazard function given input data $h(t\mid\mathbf{y}, \pmb{\eta})$ to be the product of a baseline function, which is a function of time, and a parametric function of the input data $\mathbf{y}$ and $\pmb{\eta}$. $\mathbf{y}$ and $\pmb{\eta}$ denote endogenous measurements and  exogenous measurements, respectively. Using the assumption of a linear relationship between the log-risk function and the covariates, the Cox model has the form
\begin{equation}
\label{eq:ht_Cox}
h(t\mid\mathbf{y}, \pmb{\eta})=h_0(t)\exp(\pmb{\tau}^T\mathbf{y}+\pmb{\gamma}^T\pmb{\eta}),
\end{equation}
where $h_0(t)$ is the baseline hazard function, and $\pmb{\tau}$ and $\pmb{\gamma}$
are the vectors of weights for $\mathbf{y}$ and $\pmb{\eta}$.

In BERTSurv, the log-risk function $h_{\theta}(\mathbf{x})$ is the output node from the neural network:
\begin{equation}
\label{eq:ht_BERTSurv}
h(t\mid\mathbf{y}, \pmb{\eta})=h_0(t)\exp(h_{\theta}(\mathbf{x})),
\end{equation}
where the input $\mathbf{x}$ includes $\mathbf{y}$, $\pmb{\eta}$ and clinical notes.
The likelihood function for the survival model is as follows:
\begin{equation}
\label{eq:loglikone}
p(T, \delta)=h(T)^{\delta}S(T).
\end{equation}
When $\delta=1$, it means that the event is observed at time $T$. When $\delta=0$, the event has not occurred before $T$ and it will be unknown after $T$. The time $T$ when $\delta=0$ is called the censoring time, which means the event is no longer observable.

The Cox partial likelihood is parameterized
by $\pmb{\tau}$ and $\pmb{\gamma}$ and defined as
\begin{equation}
\label{eq:pl}
{\rm {PL}}(\pmb{\tau}, \pmb{\gamma}) = \prod\limits_{i=1}^n \{\frac{\exp(\pmb{\tau}^T\mathbf{y}+\pmb{\gamma}^T\pmb{\eta})}
{\sum_{j\in R_i}\exp(\pmb{\tau}^T\mathbf{y}+\pmb{\gamma}^T\pmb{\eta})} \}
^{\Delta_i},
\end{equation}
where $\Delta_i = I(T_i \leq C_i)$. $C_i$ is the censoring time for the $i$th patient, and $I(*)$ is the indicator function. 

We use the Breslow estimator~\cite{Breslow} to estimate the cumulative baseline hazard $\widehat{H_0}(t) = \int_{0}^{t}\widehat{h_0}(u)du$:
\begin{equation}
\label{eq:h0t}
\widehat{H_0}(t)=\sum_{i = 1}^{n} \frac{I(T_i \leq t)\Delta_i}{\sum_{j\in R_i}\exp(\pmb{\tau}^T\mathbf{y}+\pmb{\gamma}^T\pmb{\eta})}.
\end{equation}

\section{Experiments and Analysis}\label{ex}
Throughout this section, we randomly pick 70\% of the trauma data as training and the rest as testing. Considering the size of our dataset and the training time, we apply 5-fold cross-validation on the trauma training dataset and grid search for hyperparameter choice. Our hyperparameters are described in Table~\ref{Hyp_tab}. Note that the sequence length and batch size choices are limited by GPU memory.

\begin{table}
\caption{Hyperparameters}\label{Hyp_tab}
\begin{center}
\begin{tabular}{|l|l|l|}
\hline
Hyperparameters  &   Survival analysis & Mortality prediction\\
\hline
Batch size  & 24 & 16\\
Sequence length &512 & 512\\
Epoch   & 4 & 4\\
Dropout rate   & 0.1 & 0.1\\
Learning rate   & 1e-2 & 4e-2\\
BERT choice &  clinical BERT & clinical BERT\\
Activation & SELU & ReLU\\

\hline
\end{tabular}
\end{center}
\end{table}

\begin{figure}[ht]
\centering
\includegraphics[width=0.6\textwidth]{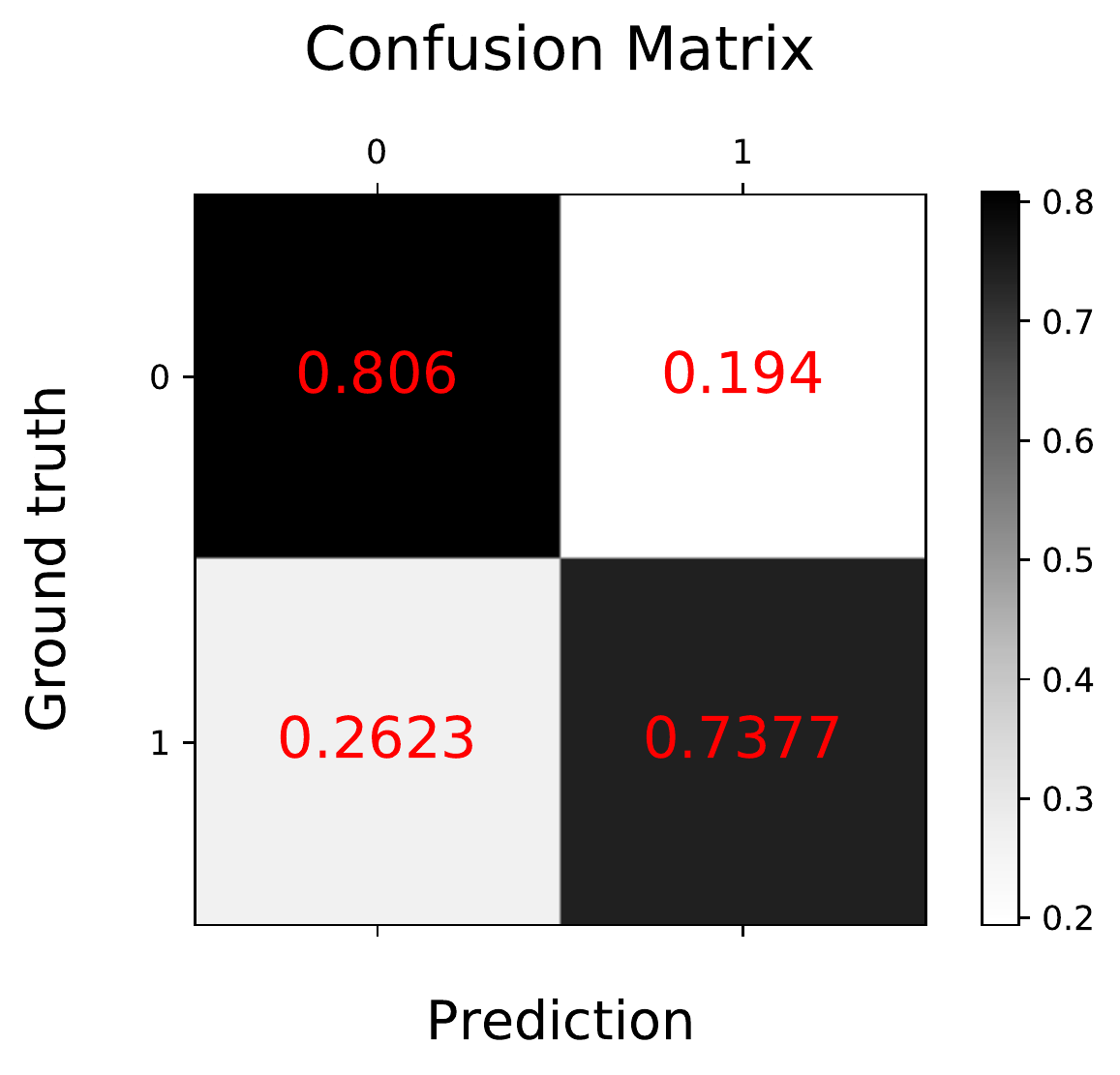}
\caption{Averaged confusion matrix for mortality prediction over 5-fold cross validation on our trauma training dataset.} \label{CV_CM}
\end{figure}
\begin{figure}[ht]
\centering
\includegraphics[width=0.6\textwidth]{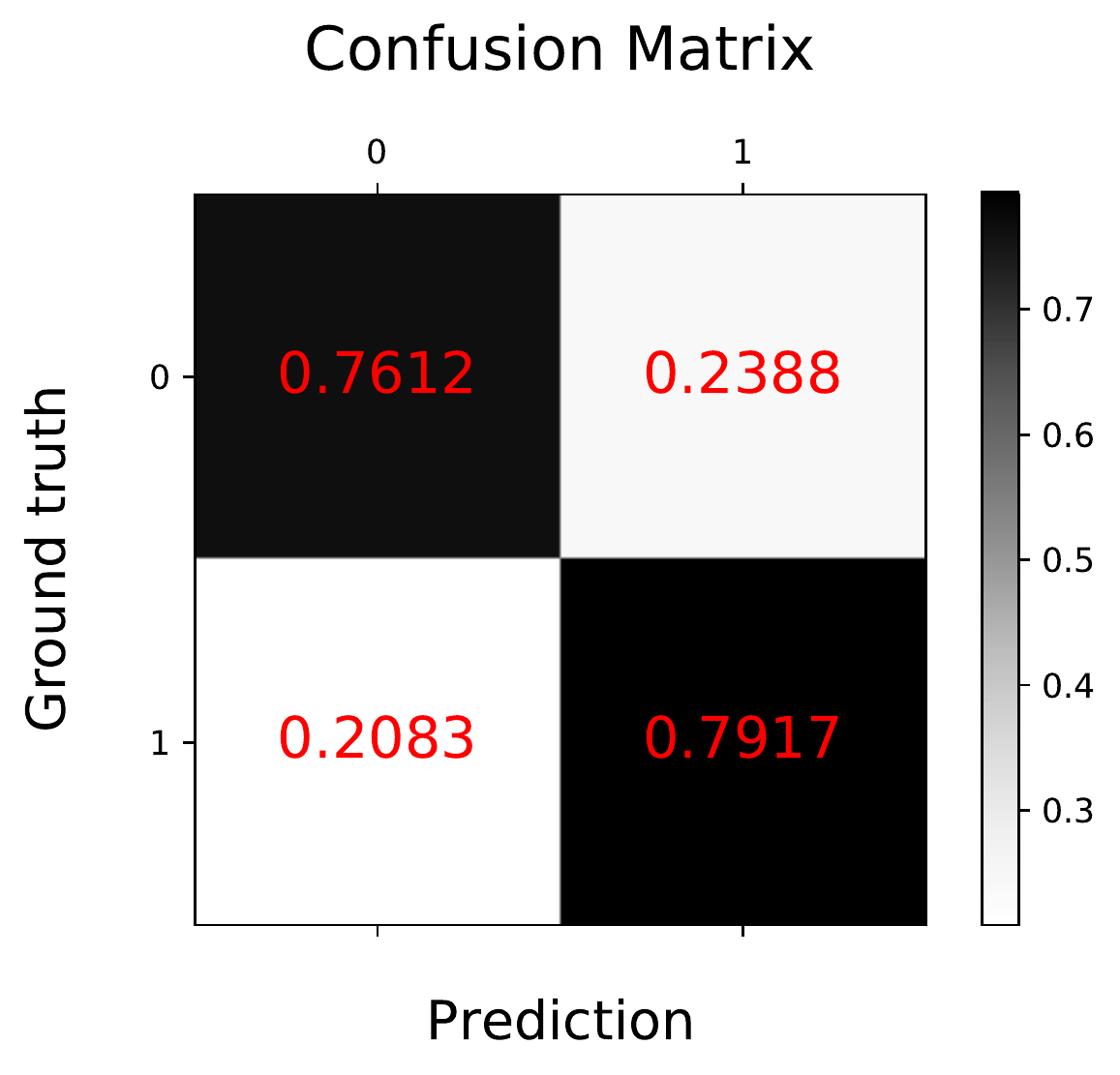}
\caption{Confusion matrix for mortality prediction on trauma testing dataset.} \label{CM}
\end{figure}
\begin{figure}[p]
\centering
\includegraphics[width=0.6\textwidth]{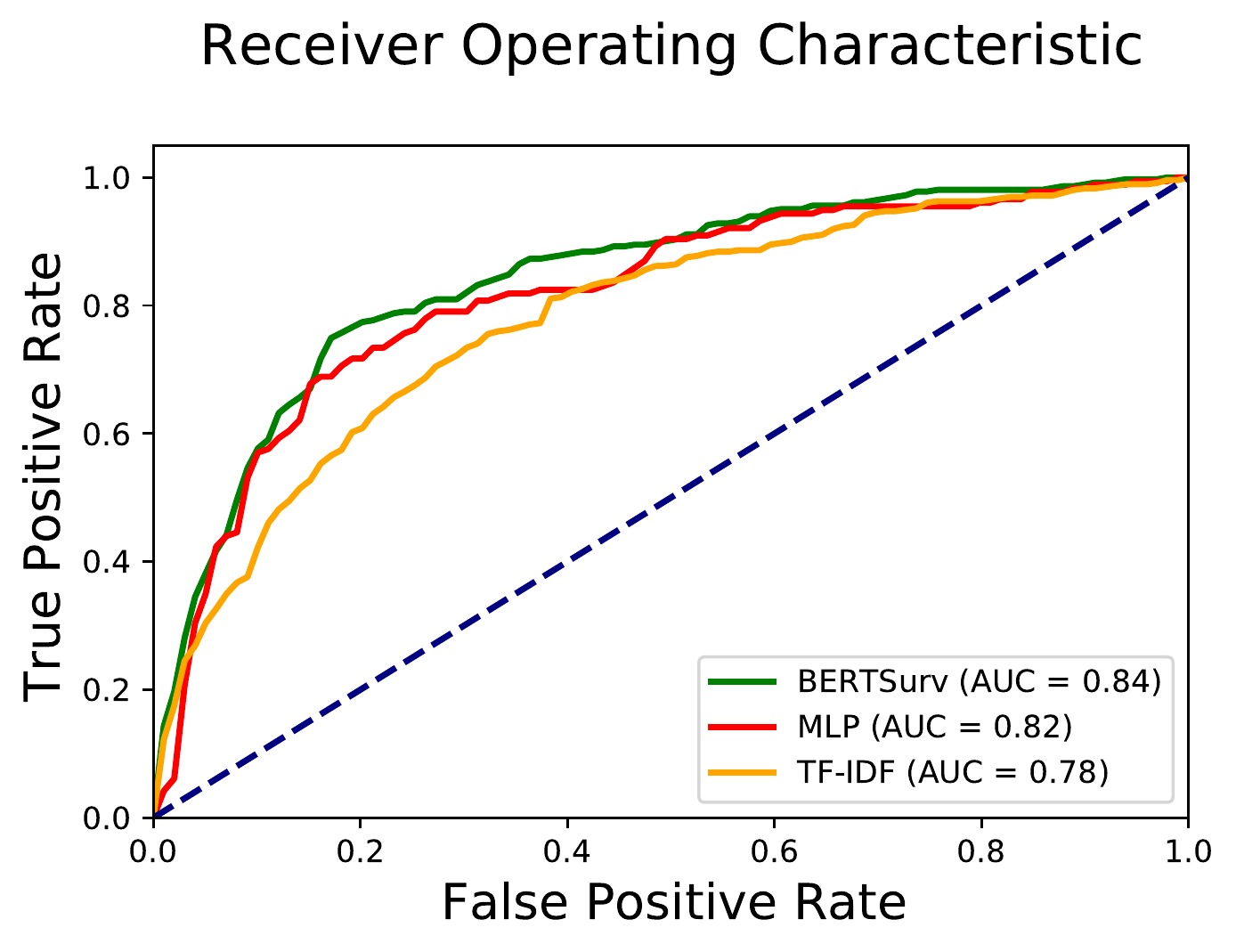}
\caption{Receiver operating characteristic (ROC) curve for mortality prediction over 5-fold cross validation on our trauma training dataset. BERTSurv outperforms both baselines.} \label{ROC_CV}
\end{figure}
\begin{figure}[p]
\centering
\includegraphics[width=0.6\textwidth]{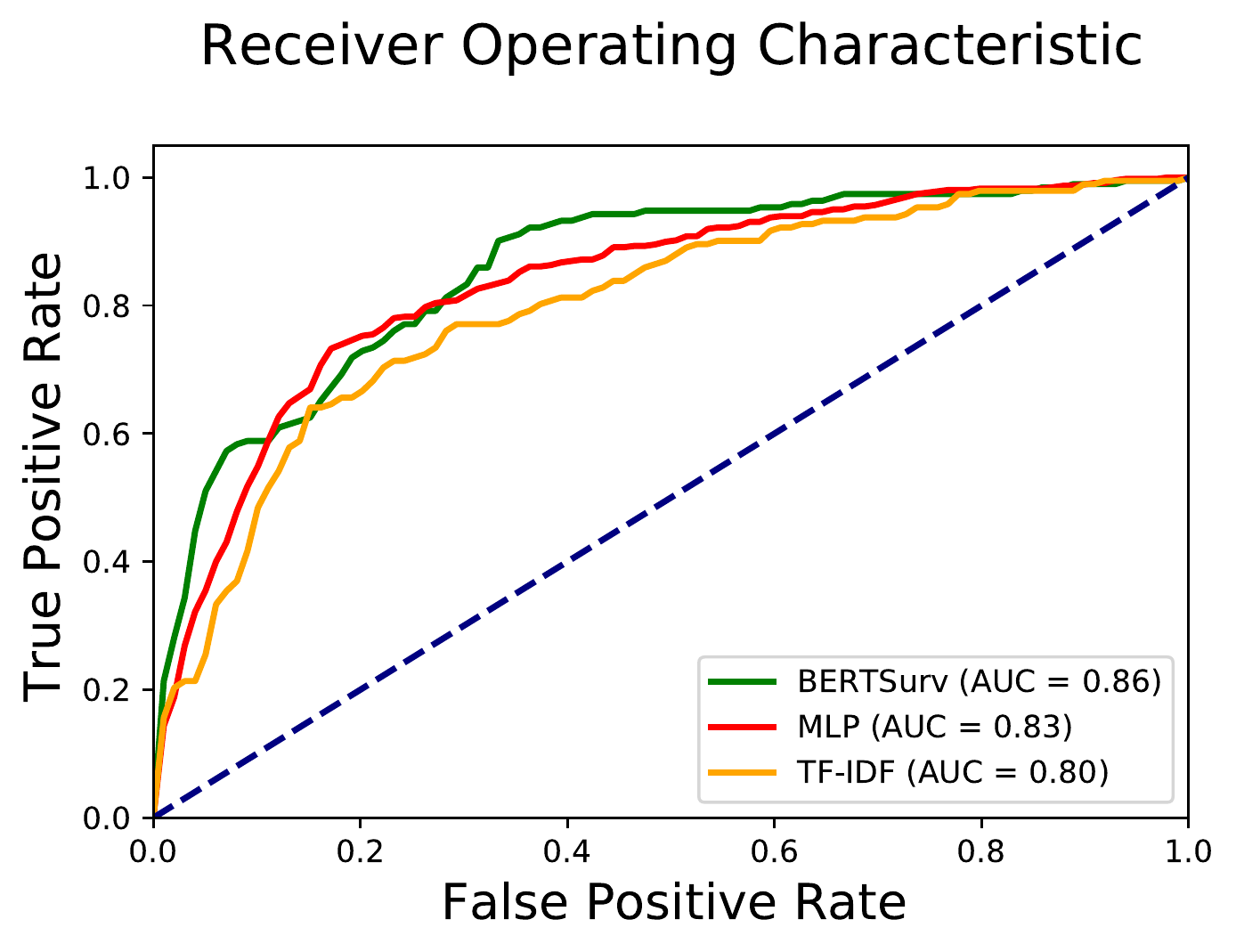}
\caption{Receiver operating characteristic (ROC) curve for mortality prediction in trauma testing dataset. BERTSurv outperforms both baselines.} \label{ROC_test}
\end{figure}

Using the clinical notes and measurements, we formulate the mortality prediction problem as a binary classification problem.  Fig.~\ref{CV_CM} shows the averaged cross validation confusion matrix for mortality prediction in the trauma training dataset. The testing confusion matrix for mortality prediction is presented in Fig.~\ref{CM}. Dominant numbers on the diagonals of both confusion matrices indicate that BERTSurv achieves high accuracy for both of the outcomes (death/discharge). With BCE loss, we apply two baselines: MLP without notes and the TF-IDF mortality model. In MLP without notes, we  consider only the measurements and build a MLP with 3 feed-forward layers for mortality outcomes. In the TF-IDF mortality model, we represent notes with TF-IDF vectors and build a support vector machine (SVM) on TF-IDF vectors combined with measurements for mortality prediction. We use AUC-ROC as our performance metric for mortality prediction, as it is commonly used in survival prediction~\cite{auc1,auc2}. AUC-ROC represents the probability that a classifier ranks the risk of a randomly chosen death patient (class $1$) higher than a randomly chosen discharged patient (class $0$). As is shown in Fig.~\ref{ROC_test}, BERTSurv achieved an AUC-ROC of 0.86, which outperforms MLP without notes by 3.6\%. BERTSurv also outperforms MLP without notes, with 5-fold cross validation as shown for our trauma training dataset in Fig.~\ref{ROC_CV}.

\begin{figure}[p]
    \centering
    \subfloat[early discharge vs. early death]{\includegraphics[width=0.5\textwidth]{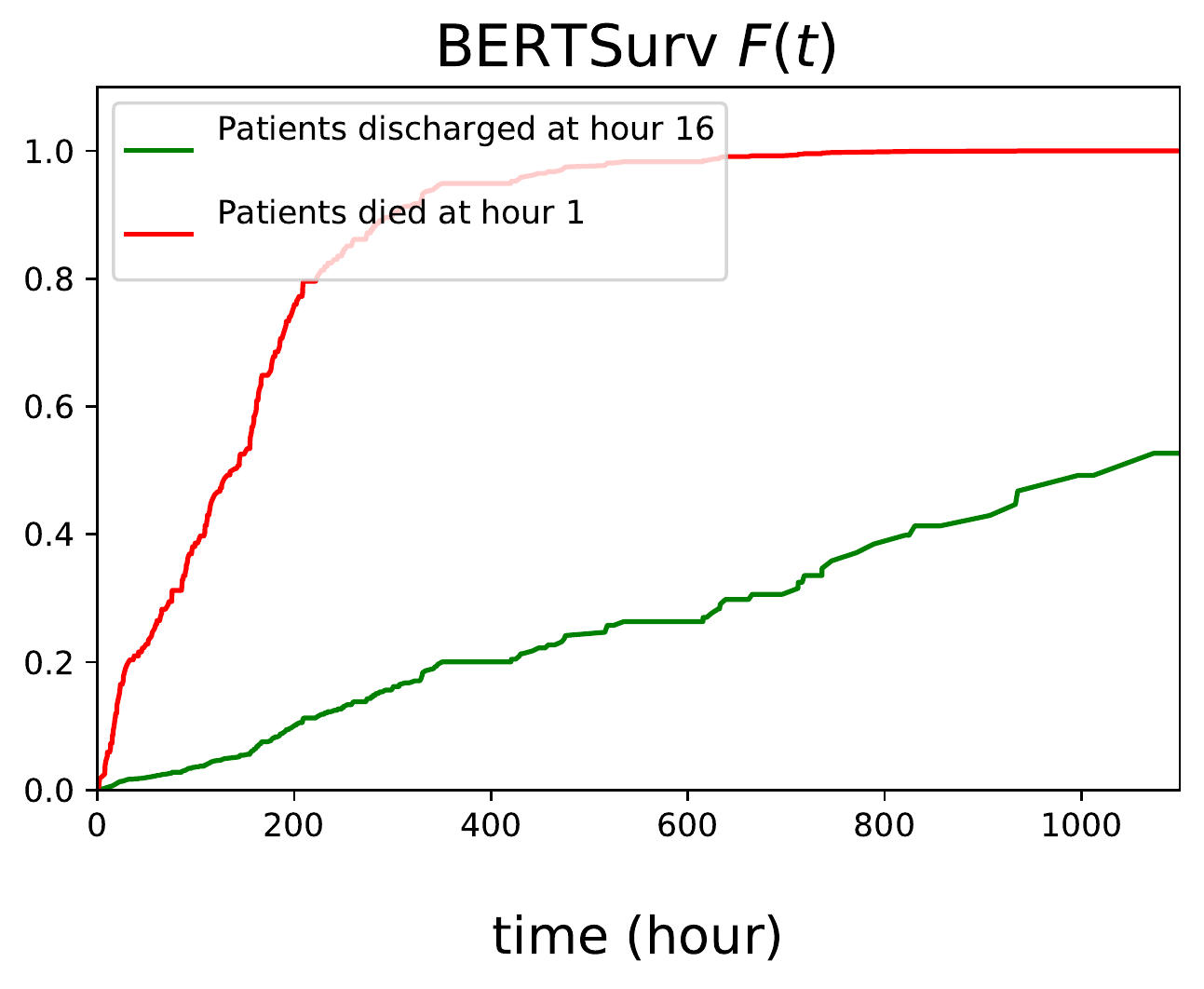}}
    \hfill
    \subfloat[early death vs. late death]{\includegraphics[width=0.5\textwidth]{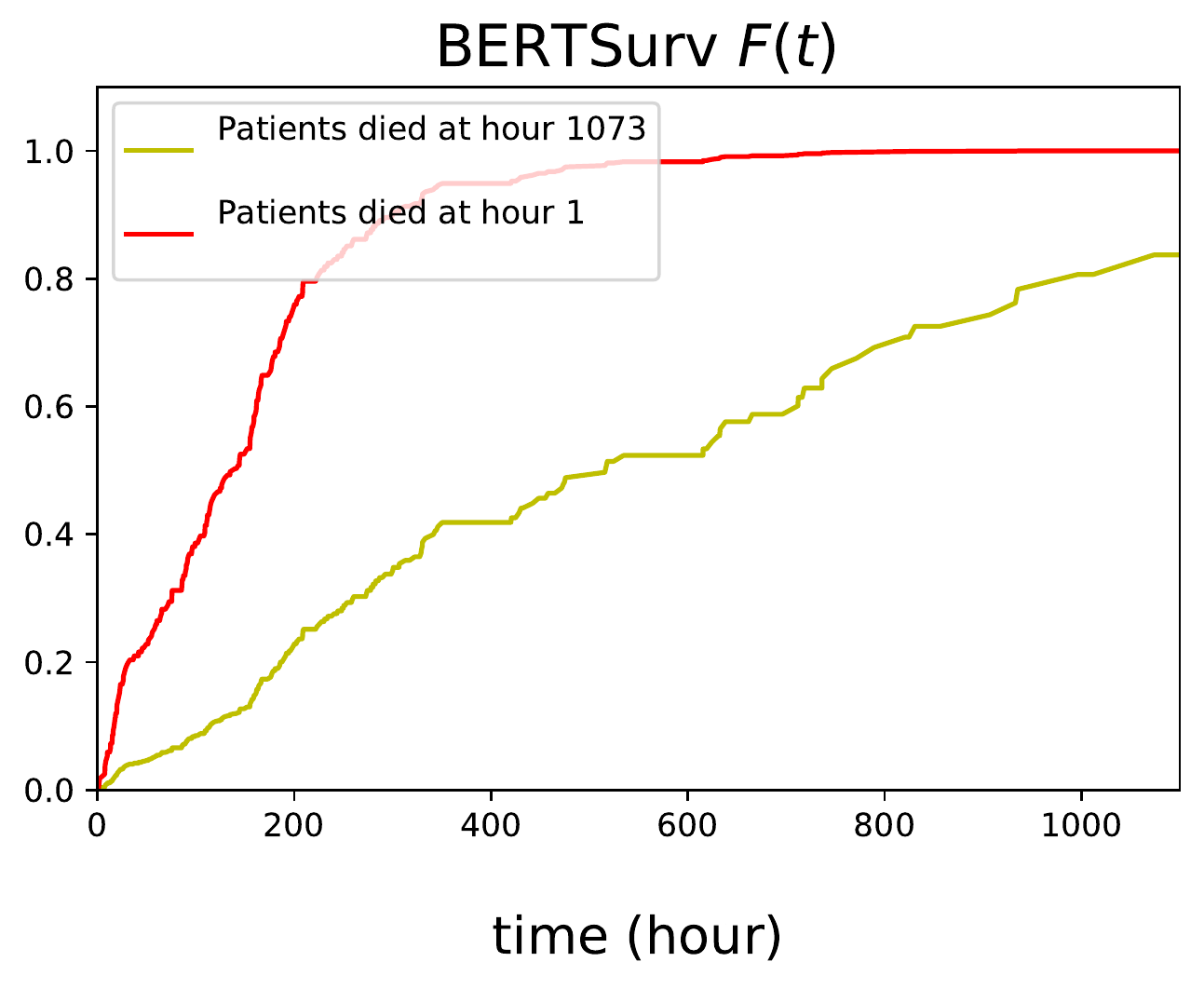}}
    \quad
    \subfloat[early discharge vs. late discharge]{\includegraphics[width=0.5\textwidth]{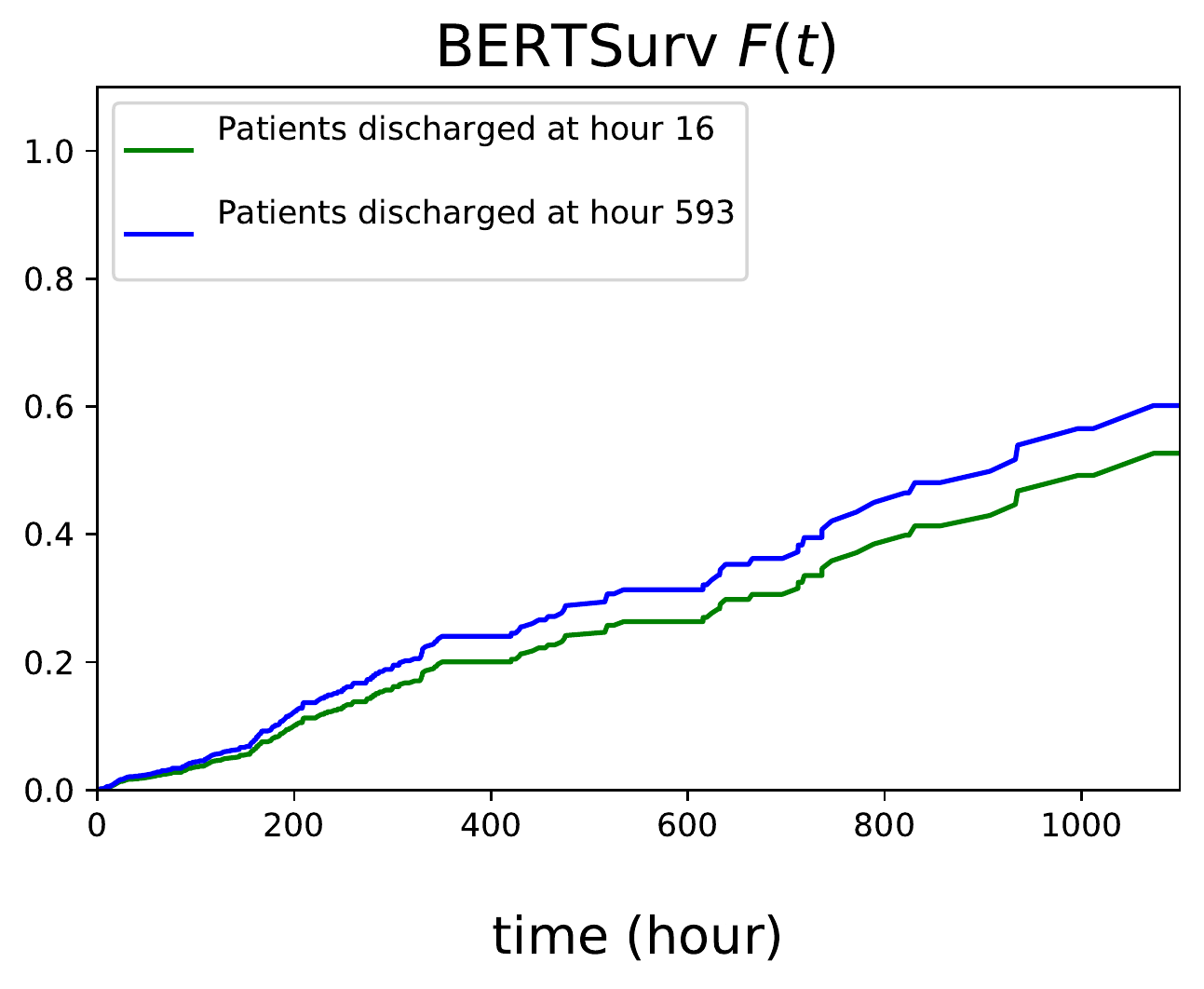}}
    \hfill
    \subfloat[late discharge vs. late death]{\includegraphics[width=0.5\textwidth]{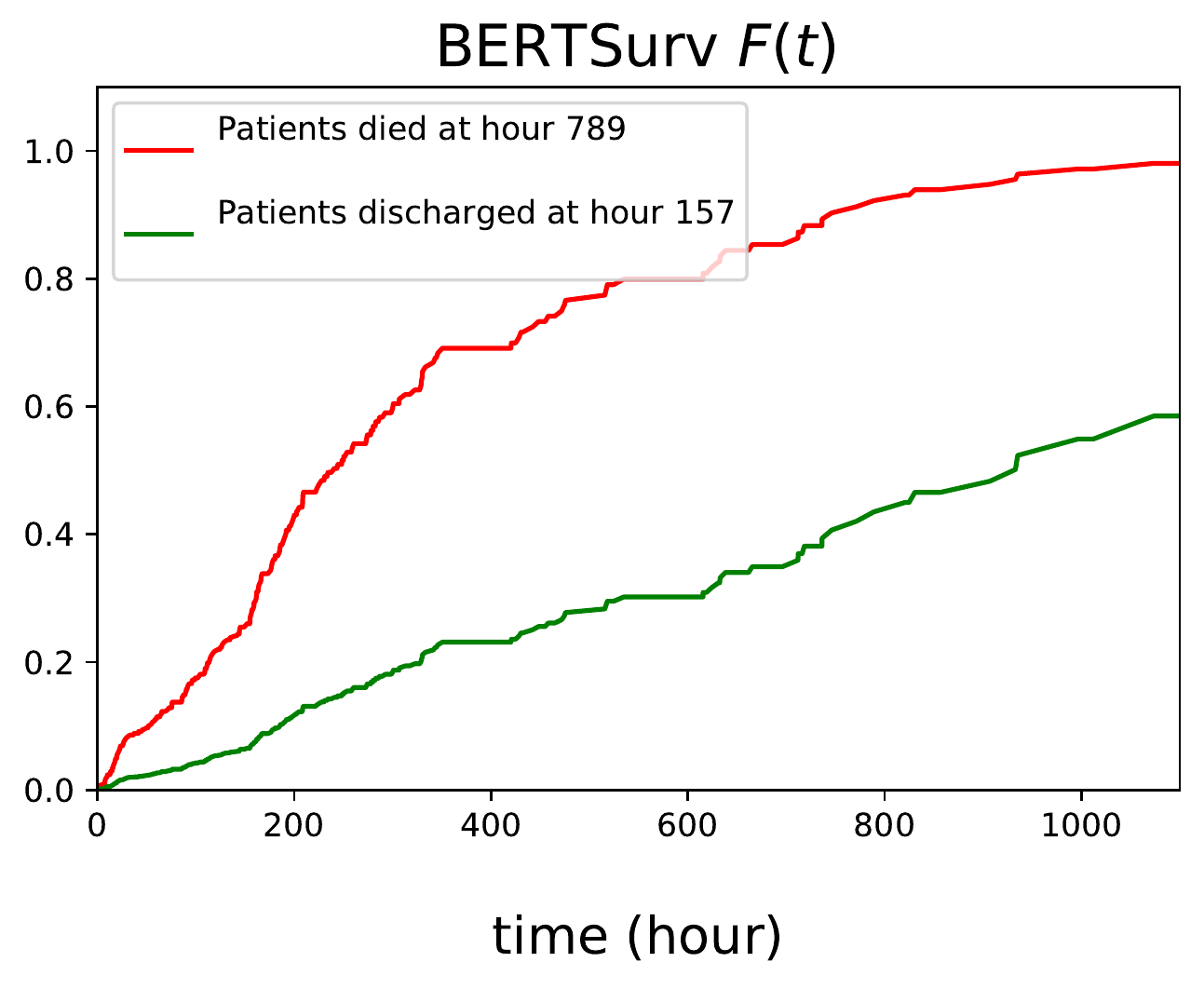}}

  \caption{Prediction of mortality as a function of time after admission to ICU using BERTSurv.}
  \label{fig:Mortality_case}
\end{figure}

\begin{figure}[p]
    \centering
    \subfloat[early discharge vs. early death]{\includegraphics[width=0.5\textwidth]{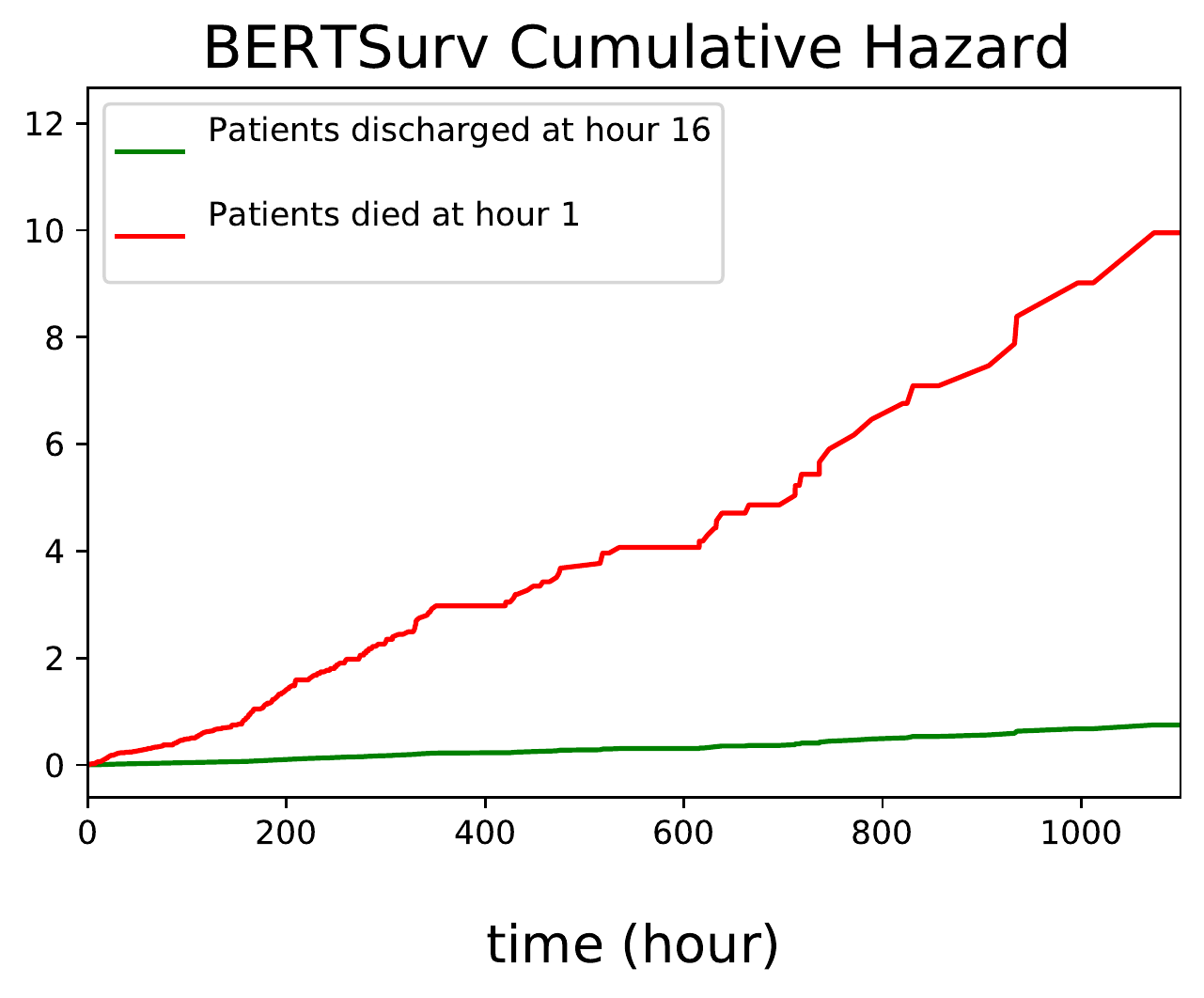}}
    \hfill
    \subfloat[early death vs. late death]{\includegraphics[width=0.5\textwidth]{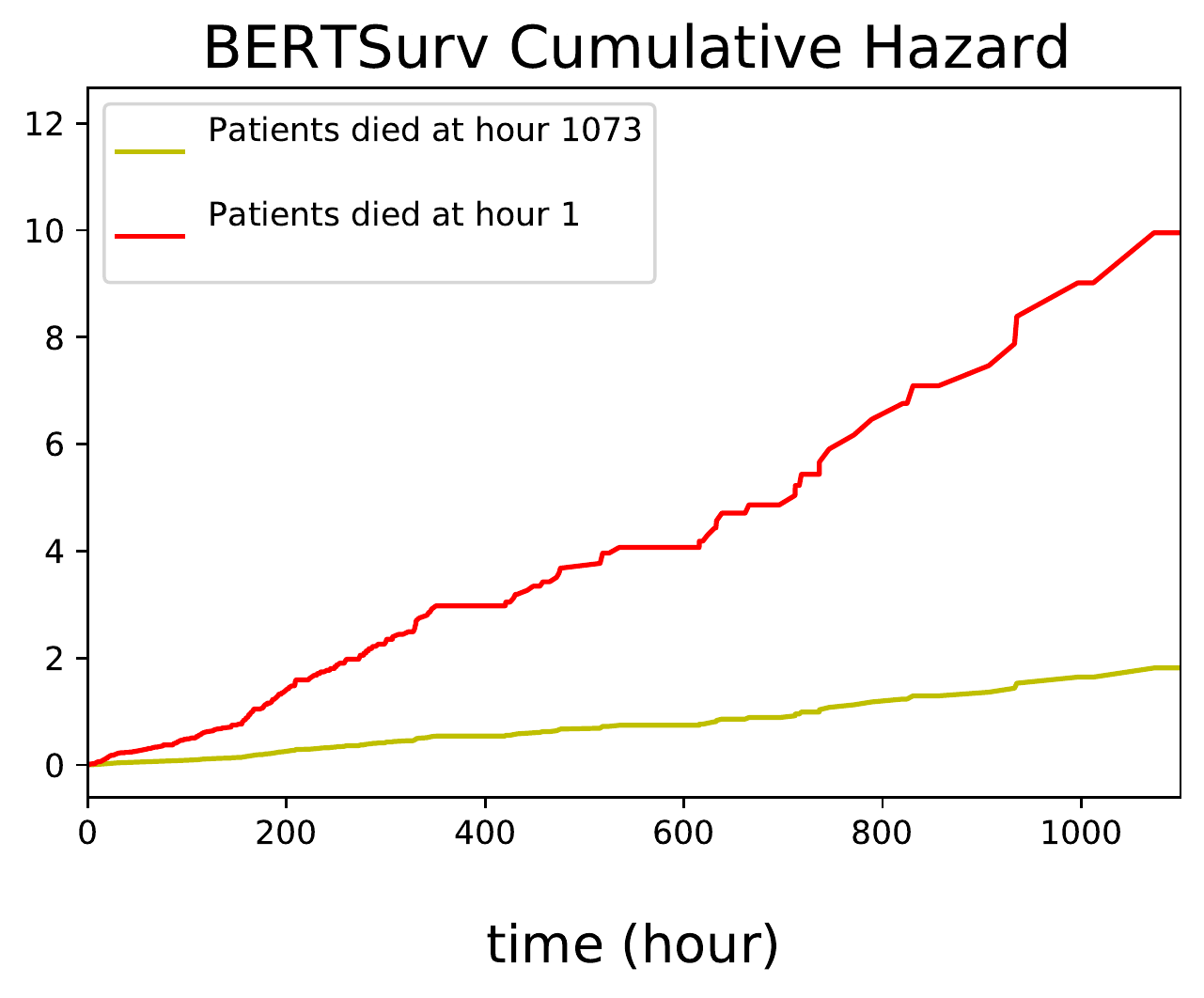}}
    \quad
    \subfloat[early discharge vs. late discharge]{\includegraphics[width=0.5\textwidth]{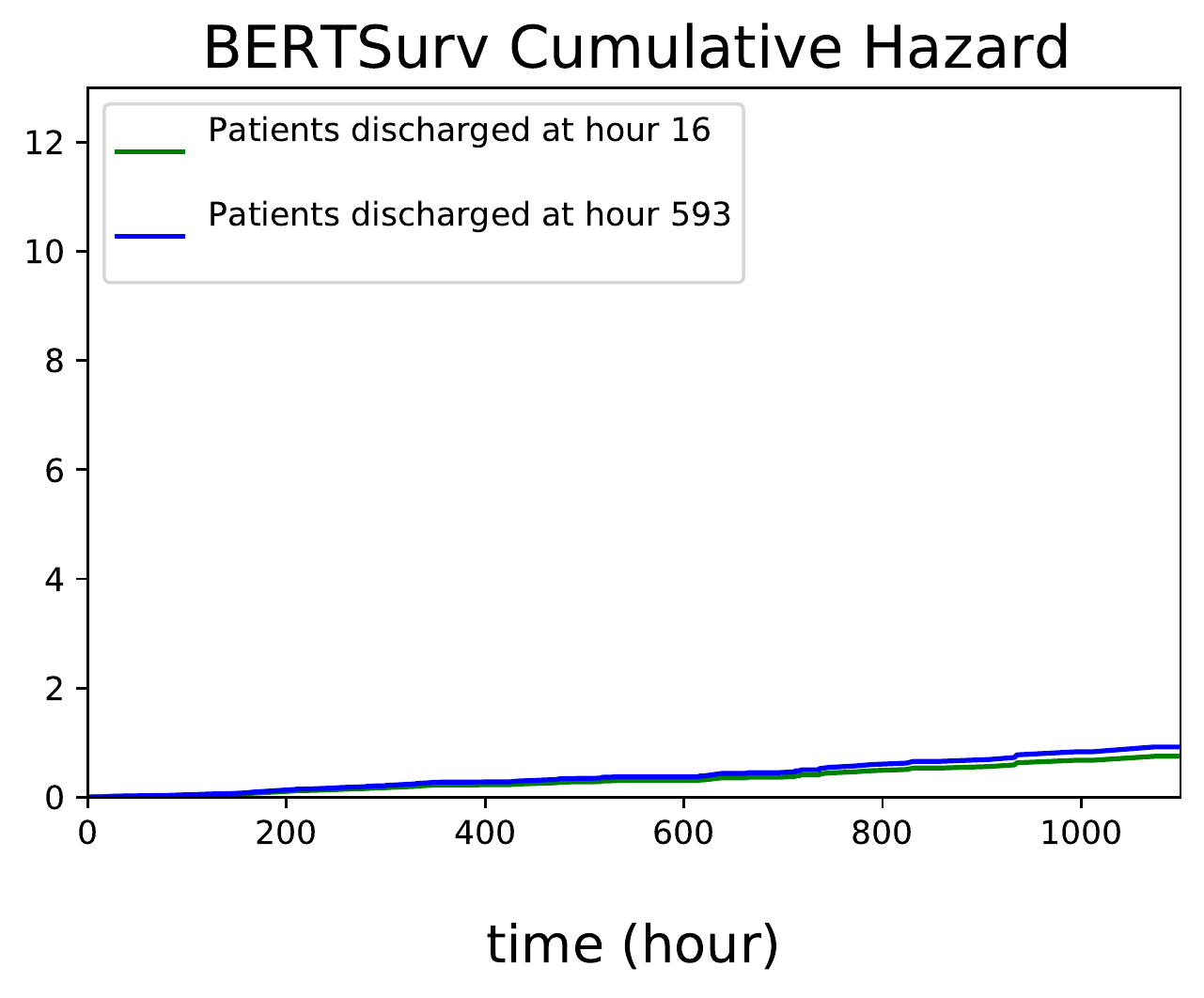}}
    \hfill
    \subfloat[late discharge vs. late death]{\includegraphics[width=0.5\textwidth]{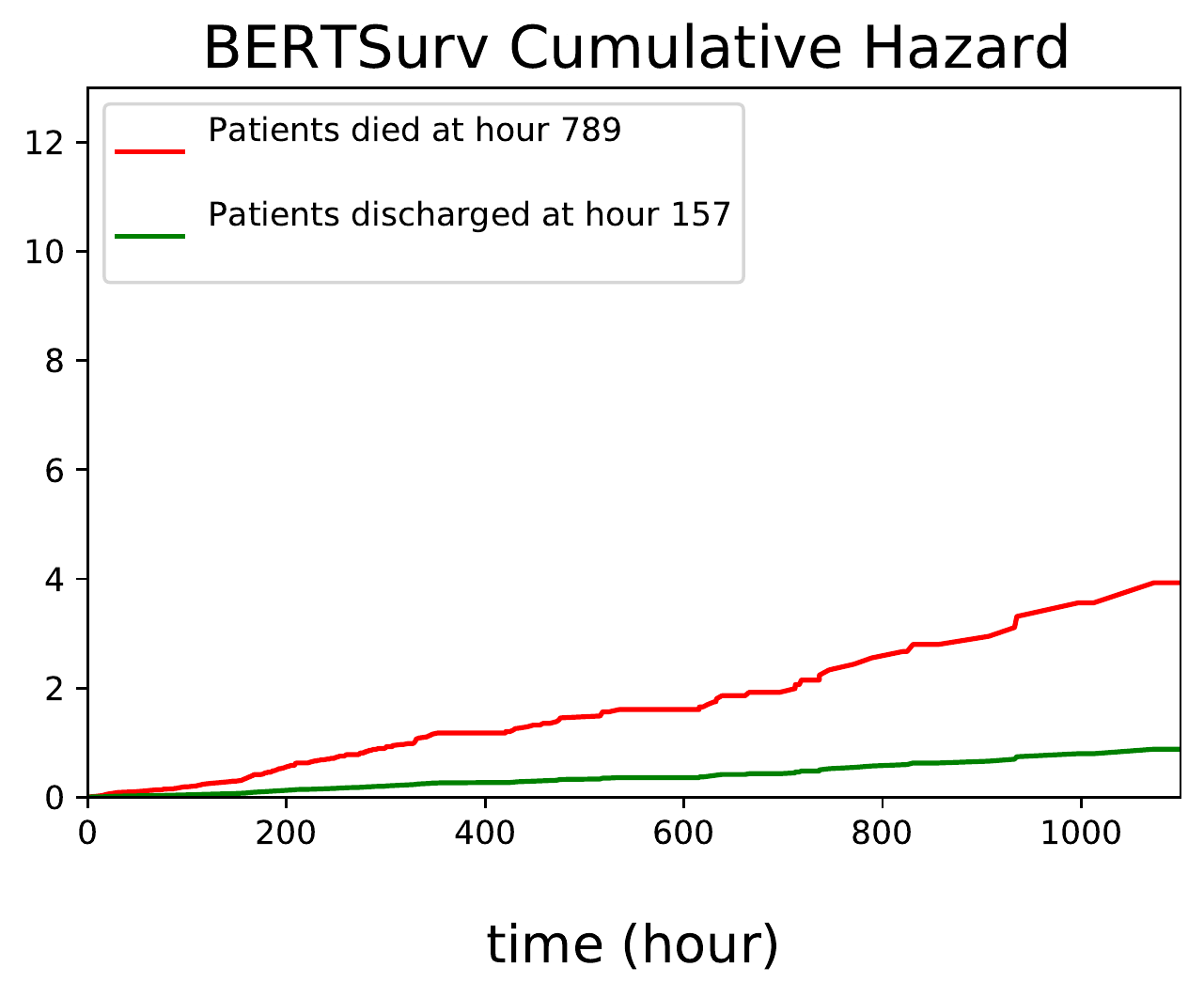}}
  \caption{Prediction of cumulative hazard function as a function of time after admission to ICU using BERTSurv. }
  \label{fig:Hazard_case}
\end{figure}

To evaluate the model's predictive performance with PLL loss on survival analysis, we measure the concordance-index
(C-index) as outlined by~\cite{cindex}. BERTSurv achieved a C-index of 0.7 on trauma patients, which outperforms a Cox model with a C-index of 0.68. To reduce the risk of ICU trauma patients  progressing to an irreversible stage, accurate and early prediction of patient condition is crucial for timely medical decisions. Mortality and cumulative hazard curves for two patients with different outcomes from BERTSurv are shown in Fig.~\ref{fig:Mortality_case} and Fig.~\ref{fig:Hazard_case}. Fig.~\ref{fig:Mortality_case}(c) 
indicates that an earlier discharged patients have a lower risk than later discharged patients, while Fig.~\ref{fig:Mortality_case}(b) shows that  patients who die early are at a relatively higher risk compared with patients who die later.
Comparing Fig.~\ref{fig:Mortality_case}(a) and Fig.~\ref{fig:Mortality_case}(d), the gap between early discharge vs. early death is larger than that of late discharge vs. late death. Similar conclusions can be drawn from the hazard curves in Fig.~\ref{fig:Hazard_case}. Such survival and hazard curves can provide physicians with comprehensive insight into patients’ condition change with time.

\begin{figure}[p]
    \centering
    \subfloat[patient died at hour 76 ]{\includegraphics[width=0.5\textwidth]{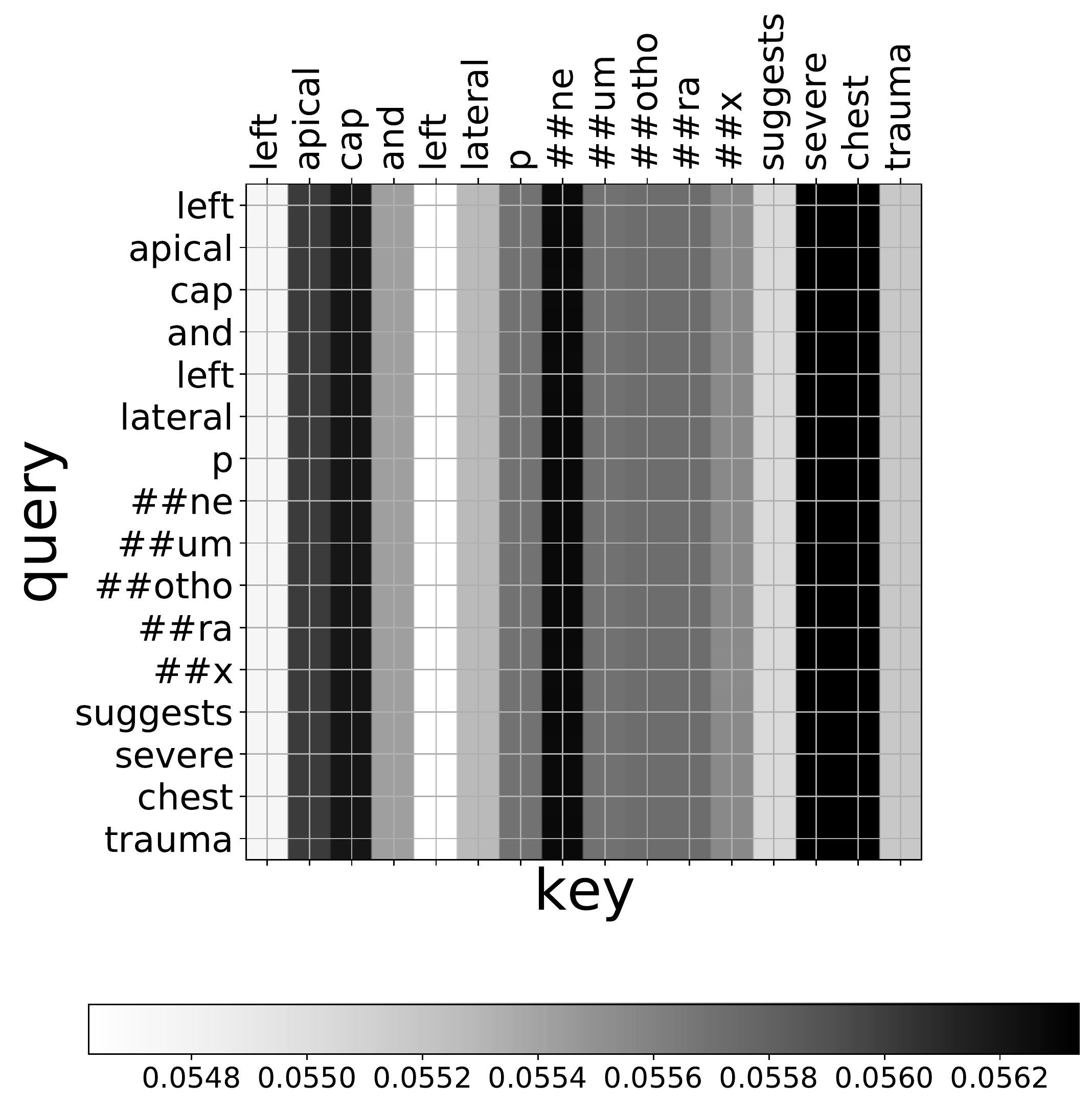}}
    \hfill
    \subfloat[patient died at hour 76]{\includegraphics[width=0.5\textwidth]{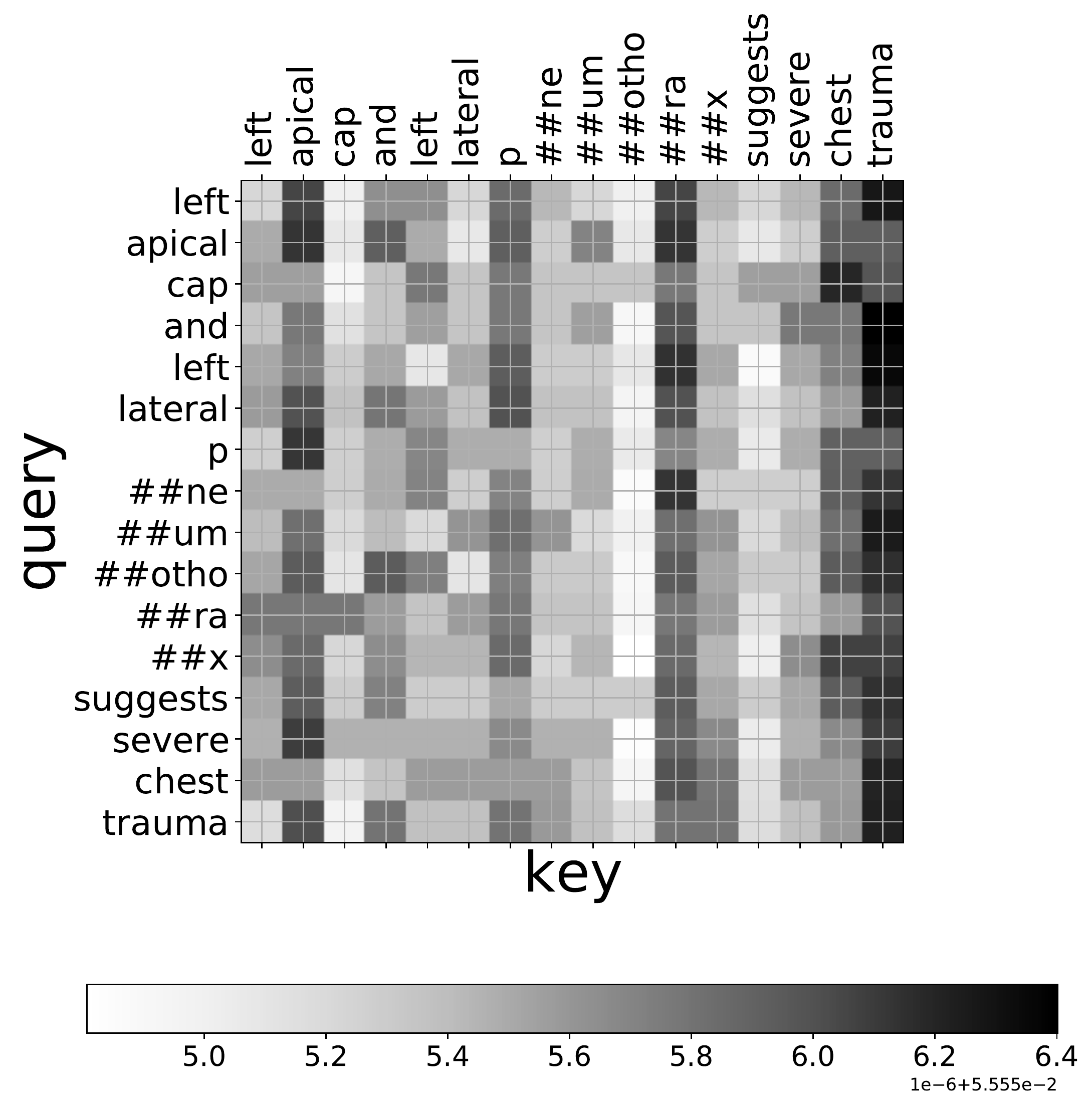}}
    \quad
    \subfloat[patient discharged at hour 85]{\includegraphics[width=0.5\textwidth]{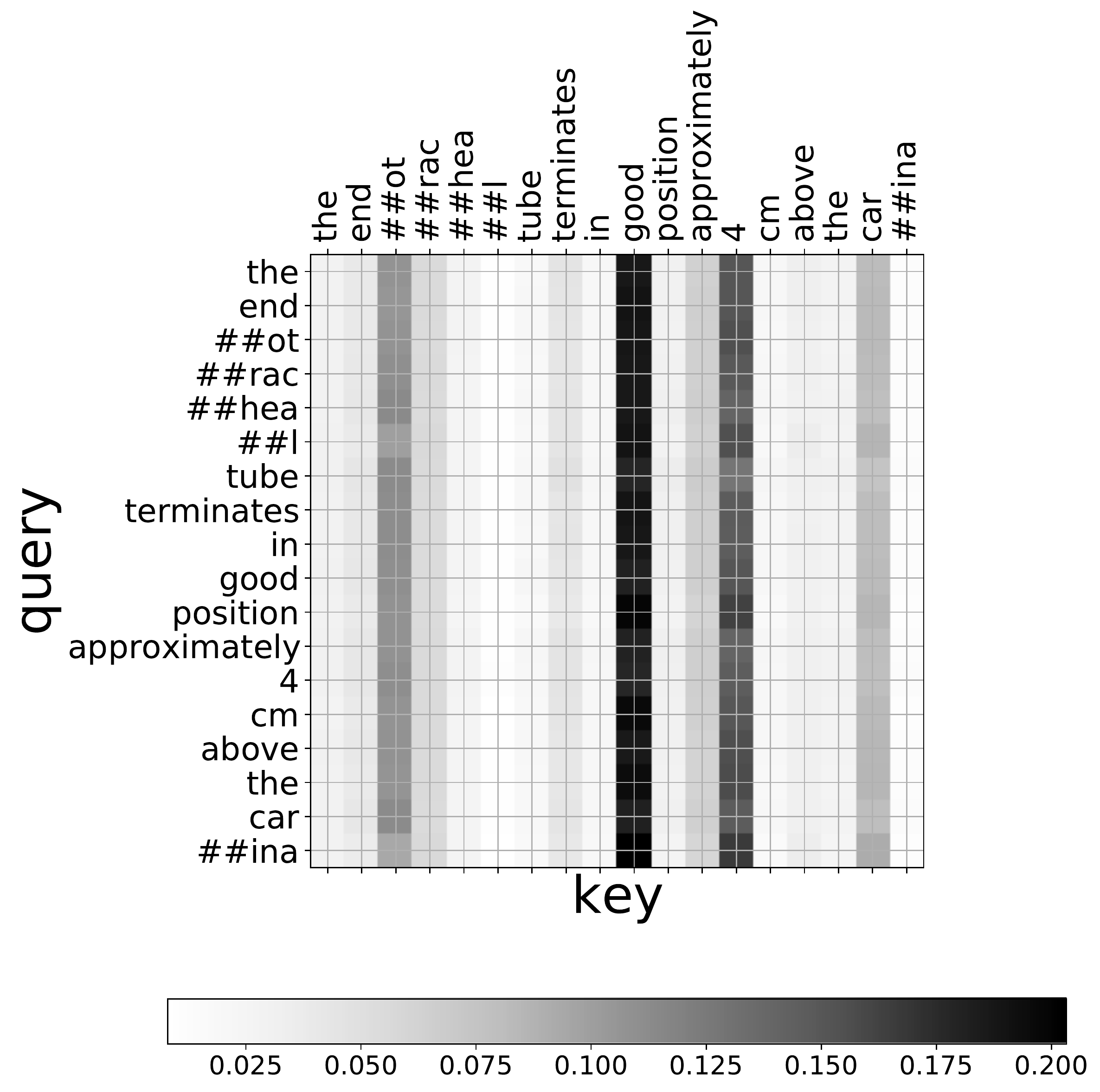}}
    \hfill
    \subfloat[patient discharged at hour 85]{\includegraphics[width=0.5\textwidth]{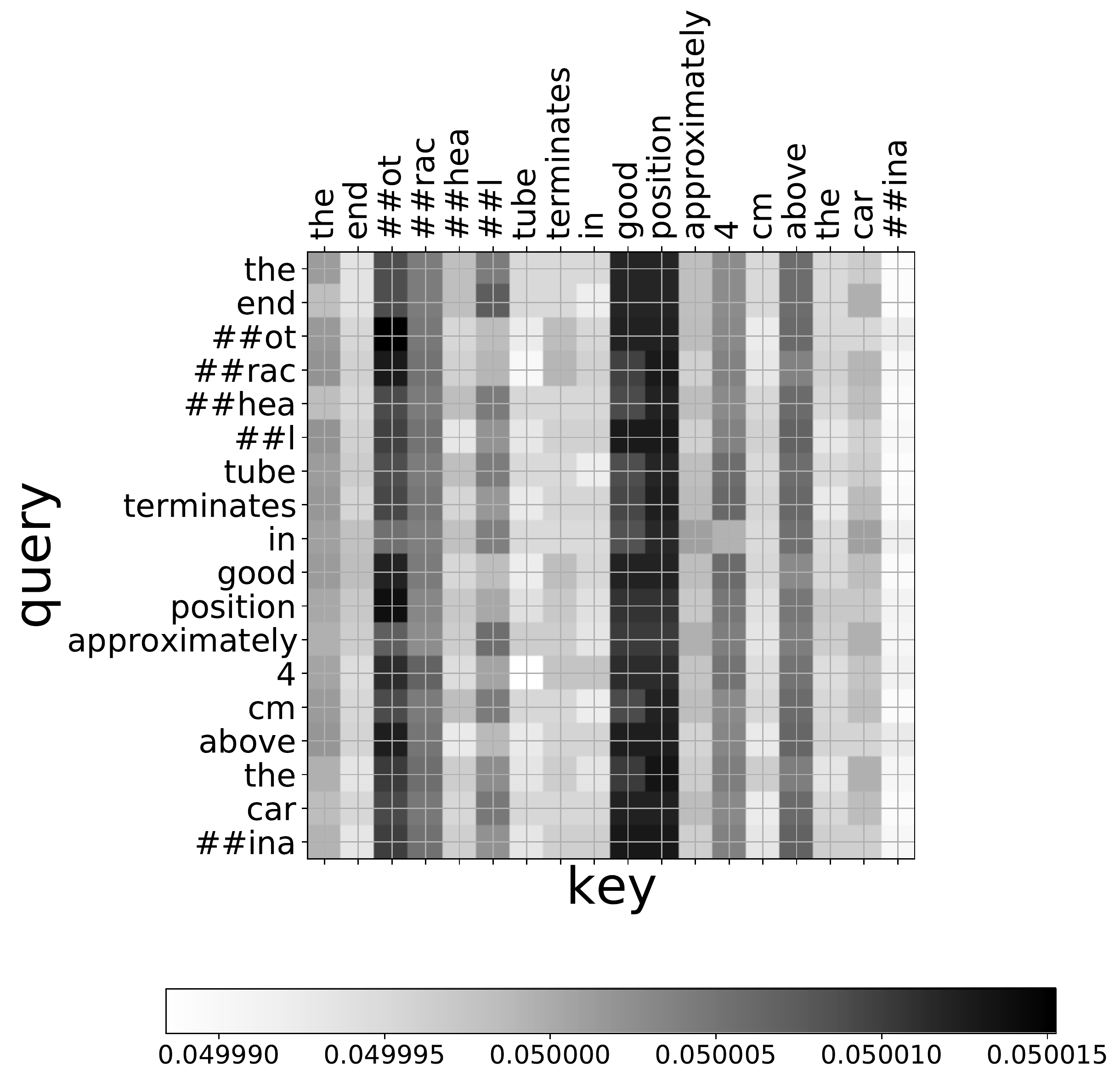}}
\caption{BERT visualization. The x-axis are the query tokens and the y-axis are the key tokens. Panels (a) and (b) are two head attention mechanisms for a patient that died at hour 76. The input notes to BERTSurv read ``left apical cap and left lateral pneumothorax suggests severe chest trauma". Panels (a) and (b) extract ``severe chest" and  ``trauma" as prominent patterns from the two heads, respectively. 
``severe chest" and ``trauma" provide insight on the patient's critically ill condition. Similarly, panels (b) and (c) are two head attention mechanisms for a patient discharged at hour 85. The input notes include ``the endotracheal tube terminates in good position approximately 4 cm above the carina". ``good” stands out in panel (c) and ``good position" emerges in panel (d). Both ``good" and ``good position" are strong indications that the patient is in a relatively benign condition.} \label{vis}
\end{figure}

Fig.~\ref{vis} depicts four self-attention mechanisms in BERTSurv which help to understand patterns in the clinical notes. In all of the panels, the x-axis represents the query tokens and the y-axis represents the key tokens. In panels (a) and (b), we analyze a clinical note ``left apical cap and left lateral pneumothorax suggests severe chest trauma
" from a patient that died at hour 76. Panels (a) and (b) are two different head attention mechanisms. Panel (a) indicates ``severe chest" and panel (b) extracts ``trauma" as prominent patterns, respectively. Similarly, panels (c) and (d) are two head attention mechanisms for a patient discharged at hour 85. The input note to BERTSurv is ``the endotracheal tube terminates in good position approximately 4 cm above the carina".  BERTSurv finds ``good" and ``good position" in panels (c) and (d), respectively. Both ``severe chest” and ``good position" help in understanding the patients' conditions and strongly correlate with the final outcomes. The indications from extracted patterns to patient outcomes show the effectiveness of BERT representation for clinical notes.

\pagebreak


\section{Discussion}\label{Dis}
We have proposed a deep learning framework based on BERT for survival analysis to include unstructured clinical notes and measurements. Our results, based on MIMIC III trauma patient data, indicate that BERTSurv outperforms the Cox model and two other baselines. We also extracted patterns in the clinical texts with attention mechanism visualization and correlated the assigned weights with survival outcomes. This paper is a proof of principle for the incorporation of clinical notes into survival analysis with deep learning models. Given the current human and financial resources allocated in preliminary clinical note analysis, 
our method has foreseeable potential to save labor costs, and further improve trauma care.
Additional data and work are needed, however, before the extent to which survival analysis can benefit from deep learning and NLP methods can be determined.

\section{Acknowledgments}
This work was funded by the National Institutes for Health (NIH) grant
NIH 7R01HL149670. We acknowledge helpful discussions from Dr. Rachael A. Callcut of the University of California, Davis.

%
%
%
%

\end{document}